\documentclass[3p,review]{elsarticle}

    \usepackage[english]{babel}
	\usepackage{dsfont}
    \usepackage{amsmath}
	\usepackage{amssymb}
	\usepackage[%
  		breaklinks=true,%
  		colorlinks=true,%
  		linkcolor=blue,anchorcolor=blue,%
 	 	citecolor=blue,filecolor=blue,%
  		menucolor=blue,%
  		urlcolor=green]{hyperref}
	\usepackage{setspace}
	\usepackage{algorithm}
	\usepackage{algpseudocode}
	\usepackage{pgfplots}
    \usepackage{amsthm}
	\usepackage{ecrc} % The `ecrc' package must be called to make the CRC functionality available

 	\newtheorem{satz}{Theorem}

	\newtheorem{lemma}[satz]{Lemma}
	
	\newproof{pf}{Proof}

	\pgfplotsset{compat=newest}

    \graphicspath{{./Img/}}

    \algrenewcommand\algorithmicindent{0.7em}

	\volume{00}

	\firstpage{1}
	
	\runauth{G.~Ehrensperger $\cdot$ A.~Ostermann $\cdot$ F.~Schwitzer}

\begin{document}
\begin{frontmatter}
    \title{Fast algorithms for morphological operations using run-length encoded binary images}
    
    \author[uibk,besi]{G.~Ehrensperger\corref{cor1}}
\ead{gregor.ehrensperger@student.uibk.ac.at}
    
    \author[uibk]{A.~Ostermann}
\ead{alexander.ostermann@uibk.ac.at}
    
    \author[besi]{F.~Schwitzer}
\ead{felix.schwitzer@besi.com}

\address[uibk]{University of Innsbruck, Department of Mathematics, Technikerstra{\ss}e 13
Innsbruck, Austria}

\address[besi]{Besi Austria GmbH, Innstra{\ss}e 16, Radfeld, Austria}

\cortext[cor1]{Corresponding author}

		\begin{abstract} \label{ch:abstract}This paper presents innovative algorithms to efficiently compute erosions and dilations of run-length encoded (RLE) binary images with arbitrary shaped structuring elements. An RLE image is given by a set of runs, where a run is a horizontal concatenation of foreground pixels. The proposed algorithms extract the skeleton of the structuring element and build distance tables of the input image, which are storing the distance to the next background pixel on the left and right hand sides. This information is then used to speed up the calculations of the erosion and dilation operator by enabling the use of techniques which allow  to skip the analysis of certain pixels whenever a hit or miss occurs. Additionally the input image gets trimmed during the preprocessing steps on the base of two primitive criteria. Experimental results show the advantages over other algorithms. The source code of our algorithms is available in \texttt{C++}.
\end{abstract}


%MSC Codes 2010:
%  	94A12   	Signal theory (characterization, reconstruction, filtering, etc.)
% 	65D18   	Computer graphics, image analysis, and computational geometry [See also 51N05, 68U05]
%	65D19   	Computational issues in computer and robotic vision
% 	68U10   	Image processing
%  	94A08   	Image processing (compression, reconstruction, etc.)
% 	68U05   	Computer graphics; computational geometry
 	
	\end{frontmatter}
%	~ \newpage ~ \newpage

	\section{Introduction}
\label{ch:Introduction}

\textit{Mathematical morphology} is a general method for the analysis of spatial structures which aims at analysing the shape and form of objects \cite{Soille2003}. In a variety of industrial computer vision applications, ranging from barcode scanning to the placement of chips in semiconductor industry, mathematical morphology is being used to process images and filter noise. We are interested in analysing \textit{binary images} since these can be represented as sets. Therefore it enables us to use set-theoretical tools to process these images.  Morphological operators which are used for noise filtering are constructed using two basic operators, namely \textit{erosion} and \textit{dilation}. In this paper we propose fast erosion and dilation algorithms on two-dimensional \textit{run-length encoded} (RLE) binary images. A \textit{run} $R = \langle lx,rx,y \rangle$ is a concatenation of pixels in horizontal direction, where $y$ indicates the $y-$coordinate of these pixels, $lx$ the $x-$coordinate of the leftmost pixel and $rx$ the $x$-coordinate of the rightmost pixel, formally $R = \langle lx, rx, y \rangle = \{ (lx,y),(lx+1,y),\ldots,(rx,y) \}$. Then
a two-dimensional binary image $Z$ can also be described by its \textit{run-length representation} $Z = \bigcup^N_{n=1}R_n$. \\
\\
Previously, several very different ideas for the fast computation of erosion and dilation have been presented, such as using the decomposition of a rectangular-shaped structuring element \cite{Breuel2007}, basing the algorithm on set-theoretical investigations \cite{Ji1989,Kim2005}, solely operating on the contours of the image \cite{Narayanan2006}, or applying methods that are similar to string-matching techniques \cite{Machado2009}. Inspired by some of the ideas introduced in \cite{Ji1989, Kim2005, Machado2009} we propose two algorithms that extend these ideas by a couple of new theorems. \\
\\
An outline of our paper is as follows: In \hyperref[ch:erosionUsingRLE]{Section~\ref{ch:erosionUsingRLE}} we provide the mathematical investigation of the erosion operator on RLE images. We come up with various ideas which can be used to formulate a fast algorithm. In \hyperref[ch:dilationUsingRLE]{Section~\ref{ch:dilationUsingRLE}} we take use of the duality between erosion and dilation to port our ideas to not only work with erosion but also with dilation. In \hyperref[sec:proposedAlgorithms]{Section~\ref{sec:proposedAlgorithms}} we propose fast algorithms based on the theorems given in the previous sections and in \hyperref[ch:runtimeanalysis]{Section~\ref{ch:runtimeanalysis}} we analyse the complexity of these algorithms. Finally we present runtime experiments and comparisons with other algorithms in \hyperref[ch:experimentalResults]{Section~\ref{ch:experimentalResults}} and our conclusion can be found in \hyperref[ch:conclusions]{Section~\ref{ch:conclusions}}.
	
	\section{Preliminaries}
\label{ch:preliminaries}

As described in the \hyperref[ch:Introduction]{Introduction}, morphological operators are constructed using the two basic operators: \textit{erosion} and \textit{dilation}.
A \textit{binary image} is given by a set of pixels. Assuming $X \subseteq \mathds{Z}^d$ and the \textit{structuring element} (SE) $B \subseteq \mathds{Z}^d$ to be binary images, these operators are defined as:
\begin{align}
    \varepsilon_B(X) &= \Big\{p \in \mathds{Z}^d \mid B_p \subseteq X \Big\},\label{def:ero} \\
    \delta_B(X) &= \Big\{p \in \mathds{Z}^d \mid (B^t)_p \cap X \neq \emptyset \Big\}\label{def:dil},
\end{align}
where $\varepsilon_B(X)$ denotes the erosion of $X$ by $B$, $\delta_B(X)$ the dilation of $X$ by $B$, $B^t$ the reflection about the origin and $B_p$ the translation of $B$ by a vector $p \in \mathds{Z}^d$.
Instead of $\varepsilon_B(X)$ and $\delta_B(X)$ we also write $X \ominus B$ and $X \oplus B$ whenever it is convenient.
A two-dimensional binary image $Z$ can also be described by its \textit{run-length representation} $Z = \bigcup^N_{n=1}R_n$,
where $N$ denotes the number of runs and $R_n$ the $n$th run of $Z$. A \textit{compact representation} of an RLE image is given when concatenated pixels are defined as a single run (not divided into several runs) and when runs are not overlapping. The proposed algorithms require compact RLE binary images as input and also return the eroded or dilated image in compact RLE representation.

	\section{Erosion Using RLE}
\label{ch:erosionUsingRLE}

In this chapter, we develop a theory to speed up the calculation of the erosion operator. This is done in five steps. In \hyperref[subsec:transSE]{Section~\ref{subsec:transSE}} we formulate a theorem which allows us to translate the structuring element before calculating the erosion such that the translated structuring element contains the origin and describe, how this improves the efficiency of our algorithm. In \hyperref[subsec:remRuns]{Section~\ref{subsec:remRuns}} we formulate a primitive criterion which allows us to remove all runs of the input image that are shorter than the longest run within the structuring element. This is achieved in linear runtime complexity before calculating the erosion. \hyperref[subsec:JMT]{Section~\ref{subsec:JMT}} on one hand summarizes the results of \cite{Machado2009} which are needed to prove the lemmata and theorems stated in Sections \hyperref[subsec:JHT]{\ref{subsec:JHT}} and \ref{subsec:skip}, on the other hand it also describes an efficient way to build the erosion transform tables and skeletons of a given image. In \hyperref[subsec:JHT]{Section~\ref{subsec:JHT}} we extend the \hyperref[Machado2009:9]{Jump-Miss Theorem} by formulating the \hyperref[Erosion:3]{Jump-Hit Theorem} and in \hyperref[subsec:skip]{Section~\ref{subsec:skip}} we give another -- easy to apply -- criterion, which makes the investigation of a big number of pixels redundant (depending on the length of the shortest run of the structuring element).

\subsection{Translating the SE such that it contains the origin} 	
\label{subsec:transSE}

    When the origin $o = (0, \ldots, 0) \in \mathds{Z}^d$ is contained in $B$ ($o \in B$) expression \eqref{def:ero} obviously reduces to
    \begin{equation}
        \varepsilon_B(X) = \Big\{p \in X \mid B_p \subseteq X \Big\}.\label{def:ero2}
    \end{equation}
    The following theorem allows us to translate $B$ by a vector $q$ such that $o \in B_q$ before calculating the erosion.
    \begin{satz}\label{eroTrans}
        Let $X, B \subseteq \mathds{Z}^d$ and $q \in \mathds{Z}^d$. Then we have that:
	 	\begin{displaymath}
	 		\varepsilon_B(X) = [\varepsilon_{B_q}(X)]_q.
	 	\end{displaymath}
    \end{satz}
    \begin{pf}
	 		Obviously
	 		\begin{align*}
	 			[\varepsilon_{B_q}(X)]_q & = \{ p \mid B_{p + q} \subseteq X \}_q  \\
	 									 & = \{ p + q \mid B_{p + q} \subseteq X \} \\
	 									 & = \{ \tilde{p} \mid B_{\tilde{p}} \subseteq X \} \tag*{\footnotesize by substitution $\tilde{p} := p + q$} \\
	 									 & = \varepsilon_B(X).
	 		\end{align*}
    \end{pf}

    Since our image $X$ is given in RLE representation, we are able to look up all $p \in X$ efficiently.

\subsection{Removing short runs before eroding}
\label{subsec:remRuns}

    Let $X = \bigcup^N_{n=1} R^X_n$ and $B = \bigcup^M_{m=1} R^B_m$ be two run-length encoded binary images in compact representation. The article \cite{Ji1989} provided an elegant method to calculate the erosion directly using the RLE representations of $B$ and $X$:
	\begin{equation}
			\varepsilon_B(X) = \bigcap\limits^M_{m=1} \bigcup\limits^N_{n=1} \varepsilon_{R^B_m}(R^X_n).\label{Kim2005:7}
	\end{equation}

    \noindent While the erosion of runs can be computed very efficiently, the intersections and unions of runs take more effort. A runtime comparison with an algorithm which is based on this method can be found in \hyperref[ch:experimentalResults]{Section~\ref{ch:experimentalResults}}.
    We are able to use this result to prove the following theorem. It allows us to remove all runs of $X$ that are shorter than the shortest run within $B$ before calculating the erosion.

	\begin{satz}\label{Erosion:2}
		Let $X = \bigcup^N_{n=1} R^X_n$ and $B = \bigcup^M_{m=1} R^B_m$ be two binary images in compact RLE representation and
\begin{alignat*}{3}
    &L_{\min}        &:= &\min\limits_{m \in \left\{1, \ldots, M\right\}} \left\{ |R^B_m| \right\} \text{ as well as} \\
    &X_{L_{\min}}    &:= &\bigcup\limits_{\substack{1 \leq n \leq N \\ |R^X_n| \geq L_{\min}}} R^X_n.
\end{alignat*}
 Then we have that:
		\begin{displaymath}
			\varepsilon_B(X) = \varepsilon_B(X_{L_{\min}}).
		\end{displaymath}
	\end{satz}
		\begin{pf}
			For all runs $R^X_n$ with $|R^X_n| < L_{\min}$ and all $m \in \left\{1, \ldots, M\right\}$ the following equality obviously holds:
            \begin{equation}
                \varepsilon_{R^B_m}(R^X_n) = \emptyset.\label{Erosion:2:a}
            \end{equation}
            In the following, let $E^n_m := \varepsilon_{R^B_m}(R^X_n)$. By applying \hyperref[Kim2005:7]{(\ref{Kim2005:7})} we get:
			\begin{align*}
					\varepsilon_B(X) & = \bigcap^M_{m=1} \bigcup^N_{n=1} E^n_m 																																												 \\
													 & = \bigcap^M_{m=1} \Bigg( \bigg[ \bigcup_{\substack{1 \leq n \leq N \\ |R^X_n| \geq L_{\min}}} E^n_m \bigg] \cup \bigg[ \bigcup\limits_{\substack{1 \leq n \leq N \\ |R^X_n| < L_{\min}}} E^n_m \bigg] \Bigg) \\
													 & = \bigcap^M_{m=1} \bigcup_{\substack{1 \leq n \leq N \\ |R^X_n| \geq L_{\min}}} E^n_m \tag*{\footnotesize using (\ref{Erosion:2:a})} \\
													 & = \varepsilon_B(X_{L_{\min}}).
			\end{align*}
		\end{pf}

\subsection{Jump-Miss Theorem}
\label{subsec:JMT}

		Let $X, A \subseteq \mathds{Z}^d$ be two binary images with $|X| < \infty$, and $|A| > 1$. Then the \textit{n-fold erosion} is given by
        \begin{displaymath}
            X \ominus_n A :=    \begin{cases}
                                 X          & \text{for } n = 0, \\
                                 (X \ominus_{n-1} A) \ominus A & \text{for } n > 0.
                                \end{cases}
        \end{displaymath}
        This allows us to introduce the \textit{erosion transform} by
			\begin{displaymath}
				f^A_X(p) :=
				\begin{cases}
					\max\limits_{n > 0} \left\{ n \mid p \in X \ominus_{n-1} A \right\} & \text{if } p \in X, \\
					0				  																										 				& \text{otherwise}.
				\end{cases}
			\end{displaymath}
    The \textit{skeleton} $S^A_B$ of $B$ with respect to $A$ (where $|A| > 1$ and $o \in A$) is the set of
    all points $p \in B$ which satisfy the following inequality (see \cite[Proposition 2.7]{Machado2009}):		
    \begin{equation}
		\max\limits_{e \in A^t} \left\{f^A_B(p + e)\right\} \leq f^A_B (p).\label{mach:2}
	\end{equation}
    According to the following theorem when checking whether a structuring element $B$ is contained in $X$ at a position $h \in \mathds{Z}^d$ it is enough to compare the values $f^A_{B_h}(s + h)$ and $f^A_X(s + h)$ at all points of the skeleton $s \in S^A_B$.
	 \begin{satz}\label{Machado2009:7} \cite[Theorem 3.3]{Machado2009}
	 Let $A, B, X \subseteq \mathds{Z}^d$, $|A|, |B|, |X| < \infty$ with $o \in A$, $|A| > 1$, and $h \in \mathds{Z}^d$. Then the following are equivalent:
		\begin{enumerate}
	 		\item $\displaystyle h \in \varepsilon_B(X)$
            \item $B_h \subseteq X$
            \item $\displaystyle \forall s \in S^A_{B}: \enspace f^A_{B_h}(h + s) \leq f^A_X(h + s)$.
		\end{enumerate}
    \end{satz}
 In practice, for most structuring elements $B$ we have that $|S^A_B| \ll |B|$. So once the erosion transform and skeleton are computed, the runtime of the actual erosion reduces to $\mathcal{O}(|X||S^A_B|)$.\label{runtime}
Furthermore it was shown that it is also possible to skip the analysis of certain pixels whenever $f^A_{B_h}(h+s) - f^A_X(h+s) > 0$. This fact is described by the \textit{Jump-Miss Theorem} which will be discussed next.

    \begin{satz}\label{Machado2009:9} \textbf{(Jump-Miss Theorem)} \cite[Theorem 3.5]{Machado2009}
	 	Let $A, B, X \subseteq \mathds{Z}^d$ with $o \in A$ and $|A|, |B|, |X| < \infty$, $|A| > 1$ as well as $s \in S^A_B$ and $h \in \mathds{Z}^d$. Then
        \begin{displaymath}
            f^A_X(h + s) < f^A_{B_h}(h + s)
        \end{displaymath}
        implies
        that there are at least
        \begin{displaymath}
            k := f^A_{B_h}(h + s) - f^A_X(h + s)
        \end{displaymath}
        points $g \in \mathds{Z}^d$ such that $B_g \nsubseteq X$.
	 	These $k$ points are given by $g_i := h + i \cdot e$ for all $i \in \{0, \ldots, k - 1\}$ and $e \in A^t$.
	 \end{satz}

Now the question arises whether the erosion transform $f^A_{B_h}$ needs to be recomputed for every given $h$. The following lemma states that it is sufficient to only determine $f^A_B$.

	 \begin{lemma}\label{Machado2009:6} \cite[Proposition 3.2]{Machado2009}
	 	Let $A, B \subseteq \mathds{Z}^d$ with $|A| > 1$, $o \in A$ and $|A|, |B| < \infty$. Then for all $h \in \mathds{Z}^d$ and $s \in S^A_B$:
		\begin{displaymath}
	 		f^A_B(s) = f^A_{B_h}(h + s).
		\end{displaymath}
    \end{lemma}

    When thinking of a convenient set $A$ to work with on run-length encoded images, one immediately comes up with $A = \{ (-1,0), (0,0) \}$ or a similarly shaped form. This set ensures fast computation of $f^A_X, f^A_B$ and allows one to determine $S^A_B$ very efficiently. With this set the erosion transform $f^A_X$ evaluates to $0$ outside of $X$, to $1$ for the left-most pixel of a given run $R^X_n \subseteq X$ and increments its value for every following pixel (in horizontal direction) within the run, such that $f^A_X$ evaluates to $|R^X_n|$ for the rightmost pixel. The following theorem proves this statement.
    \begin{satz}\label{lemma:1}
        Let $A,X \subseteq \mathds{Z}^2, |X| < \infty, h \in X$, and $A = \{(-1,0), (0,0)\}$. Then
        \begin{displaymath}
            f^A_X(h) = f^A_X(h + (-1, 0)) + 1.
        \end{displaymath}
    \end{satz}
        \begin{pf}
            We distinguish between two cases:
            \begin{enumerate}
                \item Case 1 ($h + (-1,0) \notin X$): \\
                      Then by definition we have that $f^A_X(h + (-1,0)) = 0$, $h \in X = X \ominus_0 A$ and obviously
                      \begin{align*}
                      X \ominus_1 A &= \big\{ p \mid \{(-1,0),(0,0)\}_p \subseteq X \big\} \\
                                    &= \big\{ p \mid \{(-1,0) + p,(0,0) + p\} \subseteq X \big\} \not\ni h
                      \end{align*}
                      because by assumption $((-1,0) + h) \notin X$. Thus $f^A_X(h) = 1 = f^A_X(h + (-1,0)) + 1$.
                \item Case 2 ($h + (-1,0) \in X$): \\
                      Without loss of generality let $f^A_X(h + (-1,0)) = \tilde{n}$.
                      First we prove the following implication by induction:
                      \begin{align}
                        \left( h \in X \text{ and } h + (-1,0) \in X \ominus_{\tilde{n} - 1} A \right) \ \Longrightarrow \ h \in X \ominus_{\tilde{n}} A.\label{pr:2}
                      \end{align}
                      \begin{enumerate}
                        \item Base case ($\tilde{n} = 1$): When $h, h + (-1,0) \in X$, obviously also $h \in X \ominus A = \{ p \mid \{(-1,0),(0,0)\}_p \subseteq X \} = \{ p \mid \{p + (-1,0),p\} \subseteq X \}$.
                        \item Induction step ($\tilde{n} + 1$):
                        Let $h \in X$ and $h + (-1,0) \in X \ominus_{\tilde{n}} A$. Since $o \in A$, erosion is an anti-extensive operator \cite{Soille2003}, thus also $h + (-1,0) \in X \ominus_{\tilde{n} - 1} A$ and by induction hypothesis $h \in X \ominus_{\tilde{n}} A$. We conclude
                        $X \ominus_{\tilde{n} + 1} A = \big\{ p \mid \{(p + (-1,0), p)\} \subseteq X \ominus_{\tilde{n}} A \big\} \ni h$.
                        This proves \eqref{pr:2}.
                      \end{enumerate}
                      From $f^A_X(h + (-1, 0)) = \tilde{n}$ we conclude that $h + (-1, 0) \in X \ominus_{\tilde{n} - 1} A$ and $h + (-1, 0) \notin X \ominus_{\tilde{n}} A$. Thus
                      \begin{displaymath}
                        X \ominus_{\tilde{n} + 1} A = \big\{ p \mid \{p + (-1,0), p\} \subseteq X \ominus_{\tilde{n}} A \big\} \not\ni h
                      \end{displaymath}
                      and using \eqref{pr:2}, we get $h \in X \ominus_{\tilde{n}} A$. Putting all together, this implies $f^A_X(h) = \tilde{n} + 1$. \hfill \qedhere
            \end{enumerate}
        \end{pf}

     Note that \eqref{mach:2} and \hyperref[lemma:1]{Theorem~\ref{lemma:1}} imply that the skeleton can be retrieved by taking the rightmost pixel of each run.

\subsection{Jump-Hit Theorem}
\label{subsec:JHT}

In the previous section we described a criterion which allows us to skip the analysis of certain pixels whenever the algorithm finds a pixel $h$ such that $h \notin \varepsilon_B(X)$. We also call this a \textit{miss}. Now we formulate a criterion which applies for \textit{hits}. A hit denotes the occurrence of an $h$ such that $h \in \varepsilon_B(X)$.

	\begin{lemma}\label{Erosion:1}
		Let $X \subseteq \mathds{Z}^2$ be a binary image, $|X| < \infty$, $p \in X$ and $A = \left\{(-1,0), (0,0)\right\}$. Then the condition $f^{A^t}_X(p) = n$ implies 
		\begin{displaymath}
			\forall i \in \left\{0, \ldots, n - 1\right\}: f^A_X(p + (i,0)) = f^A_X(p) + i.
		\end{displaymath}
	\end{lemma}		
	\begin{pf}
			Obviously this statement holds for $i = 0$. Let $i \in \left\{1, \ldots, n-1\right\}$.
			For the given set $A$ the equality $f^{A^t}_X(p) = n$ implies that for all $i \in \left\{1, \ldots, n - 1\right\}$ also $(p + (i,0)) \in X$. By applying \hyperref[lemma:1]{Theorem~\ref{lemma:1}} multiple times we get $f^A_X(p + (i, 0)) = f^A_X(p) + i$.
	\end{pf}

    Note that \hyperref[Erosion:1]{Lemma~\ref{Erosion:1}} does not hold for arbitrary sets $A$.
		
	\begin{satz}
		\textbf{(Jump-Hit Theorem)}\label{Erosion:3}
		Let $B, X \subseteq \mathds{Z}^2$, $A = \left\{(-1,0), (0,0)\right\}$ and $(h_x, h_y) = h \in \mathds{Z}^2$. If for all $s \in S^A_{B}$ we have that
        $f^A_{B_h}(h + s) \leq f^A_X(h + s)$, then the complete eroded run is given by
        \begin{displaymath}
            \langle h_x, h_x + n - 1, h_y \rangle \subseteq \varepsilon_B(X),
		\end{displaymath}
        where $n := \min\limits_{s \in S^A_B} \left\{f^{A^t}_X(h + s)\right\}$.
	\end{satz}
		\begin{pf}
			Using \hyperref[Machado2009:7]{Theorem~\ref{Machado2009:7}} we get $h \in \varepsilon_B(X)$ and \hyperref[Erosion:1]{Lemma~\ref{Erosion:1}} implies for all $s \in S^A_{B}$ and all $i \in \left\{0, \ldots, n - 1\right\}$ that
			\begin{align*}
				f^A_X(h + s + (i,0)) & = f^A_X(h + s) + i \tag*{\footnotesize with \hyperref[Erosion:1]{Lemma~\ref{Erosion:1}}} \\
														 & \geq f^A_X(h + s) \tag*{\footnotesize $i \geq 0$}																			 \\
														 & \geq f^A_{B_h}(h + s), \tag*{\footnotesize by assumption}
			\end{align*}
			from which we conclude -- again with
			\hyperref[Machado2009:7]{Theorem~\ref{Machado2009:7}} -- that
			$(h + (i,0)) \in \varepsilon_B(X)$ for all $i \in \left\{0, \ldots, n-1\right\}$ and therefore $\left<h_x, h_x + n - 1, h_y\right> \subseteq \varepsilon_B(X)$.
		\end{pf}

    We want to emphasize that, in case of a miss, the length of a jump is limited by the width of the structuring element. When having a hit, however, the length of the jump is always maximal and determined by just checking the values of $f^{A^t}_X(h + s)$ at its starting pixel $h$ along the pixels $s$ of the skeleton $S^A_B$.

\subsection{Skipping the analysis of additional pixels}
\label{subsec:skip}

    The goal of this section is to find a binary image $X_{\operatorname{cut}} \subseteq X_{L_{\min}}$ such that $|X_{\operatorname{cut}}| \ll |X_{L_{\min}}|$ and
    \begin{displaymath}
        \varepsilon_B(X) = \Big\{p \in X_{\operatorname{cut}} \mid B_p \subseteq X_{L_{\min}} \Big\}.
    \end{displaymath}

    In the upcoming lemma we investigate a very special class of structuring elements. These elements include a run whose rightmost element is the origin. In a next step we are going to extend the lemma such that this condition is no longer needed.

    \begin{lemma}\label{Xcutlemma}
		Let $X = \bigcup^N_{n=1} R^X_n$ and $B = \bigcup^M_{m=1} R^B_m$ be two run-length encoded binary images in compact representation, $\hat{m} \in \{1, \ldots, M\}$ such that $R^B_{\hat{m}} = \langle lx_{\hat{m}}, 0, 0 \rangle$, and $\widetilde{X} := \bigcup^N_{n=1} \tilde{R}^X_n$, where
\begin{displaymath}
    \tilde{R}^X_n := \begin{cases} \langle lx_n + (|R^B_{\hat{m}}| - 1), rx_n, y_n \rangle & \text{if } |R^B_{\hat{m}}| \leq |R^X_n|, \\
\emptyset & \text{else} \end{cases}
\end{displaymath}
for all $n = 1, \ldots, N$.
Then we have that:
        \begin{displaymath}
            \varepsilon_B(X) = \Big\{p \in \widetilde{X} \mid B_p \subseteq X_{L_{\min}} \Big\},
        \end{displaymath}
        where $X_{L_{\min}}$ is defined as in \hyperref[Erosion:2]{Theorem~\ref{Erosion:2}}.
    \end{lemma}
        \begin{pf}
            Let $R^X_n$ be an arbitrary run of $X$, $A := \{ (-1,0), (0,0) \}$, and $h \in R^X_n$. Then, since $R^B_{\hat{m}} = \langle lx_{\hat{m}},0,0 \rangle$, using Theorems~\hyperref[eroTrans]{\ref{eroTrans}} and \hyperref[lemma:1]{\ref{lemma:1}}, the following equality holds:
            \begin{displaymath}
                f^A_{B_h}(h + (0,0)) = f^A_{B_h}(h) = f^A_B((0,0)) = |R^B_{\hat{m}}|.
            \end{displaymath}
            We distinguish between two cases:
            \begin{enumerate}
                \item Case 1: Let $|R^B_{\hat{m}}| > |R^X_n|$. Then, by using the previous equality and \hyperref[lemma:1]{Theorem~\ref{lemma:1}}, we get:
                    \begin{displaymath}
                        f^A_{B_h}(h + (0,0)) = |R^B_{\hat{m}}|  > |R^X_n| \geq f^A_X(h).
                    \end{displaymath} \hyperref[Machado2009:7]{Theorem~\ref{Machado2009:7}} implies $h \notin \varepsilon_B(X)$ for all $h \in R^X_n$.
                \item Case 2: Let $|R^B_{\hat{m}}| \leq |R^X_n|$. Using
                \hyperref[lemma:1]{Theorem~{\ref{lemma:1}}} we conclude that for the first $|R^B_{\hat{m}}| - 1$ pixels $h$ of run $R^X_n$ we have that $f^A_X(h) < |R^B_{\hat{m}}|$. This leads to:
                    \begin{displaymath}
                        f^A_{B_h}(h + (0,0)) = f^A_{B}((0,0)) = |R^B_{\hat{m}}|
                        > f^A_X(h)
                    \end{displaymath}
                    for $h \in \langle lx_n, lx_n + |R^B_{\hat{m}}| - 2, y_n \rangle$.
                    Again \hyperref[Machado2009:7]{Theorem~\ref{Machado2009:7}} implies $h \notin \varepsilon_B(X)$ for all $h \in \langle lx_n, lx_n + |R^B_{\hat{m}}| - 2, y_n \rangle$.
              \end{enumerate}
              Since $o \in B$ by assumption, \eqref{def:ero2} and \hyperref[Erosion:2]{Theorem~\ref{Erosion:2}} state that $\varepsilon_B(X) = \{ p \in X_{L_{\min}} \mid B_p \subseteq X_{L_{\min}} \}$. Additionally we just proved that for all $n \in \{1, \ldots, N\}$ and $h \in \langle lx_n, lx_n + |R^B_{\hat{m}}| - 2, y_n \rangle$ where $\langle a, b, c \rangle := \emptyset$ for $a > b$ also $h \notin \varepsilon_B(X)$. Thus the statement holds.
        \end{pf}

    Obviously it makes sense to translate $B$ by a vector $q$ such that the rightmost pixel of the longest run within $B$ is placed at the origin. This translation is described in the next theorem.

    \begin{satz}\label{Xcuttheorem}
		Let $X = \bigcup^N_{n=1} R^X_n$ and $B = \bigcup^M_{m=1} R^B_m$ be two run-length encoded binary images in compact representation and $X_{\operatorname{cut}} := \bigcup^N_{n=1} \tilde{R}^X_n$, where $L_{\max}^B := \max\limits_{m \in \{1, \ldots, M\}} \{|R^B_m|\}$ and
\begin{displaymath}
    \tilde{R}^X_n = \begin{cases} \langle lx_n + (L_{\max}^B - 1), rx_n, y_n \rangle & \text{if } L_{\max}^B \leq |R^X_n|, \\
\emptyset & \text{else} \end{cases}
\end{displaymath}
for all $n = 1, \ldots, N$. Let $R^B_{\hat{m}} \subseteq B$ be any run such that $|R^B_{\hat{m}}| = L^{B}_{\max}$. Then $q := (rx^B_{\hat{m}}, y^B_{\hat{m}})$ denotes the rightmost pixel of the longest run $R^B_{\hat{m}}$ within $B$ and we have that:
        \begin{align*}
            \varepsilon_B(X) = \left[\varepsilon_{B_{-q}}(X)\right]_{-q} = \big\{p \in X_{\operatorname{cut}} \mid B_{p-q} \subseteq X_{L_{\min}} \big\}_{-q},
        \end{align*}
        where $X_{L_{\min}}$ is defined as in \hyperref[Erosion:2]{Theorem~\ref{Erosion:2}}.
    \end{satz}
        \begin{pf}
            \hyperref[eroTrans]{Theorem~\ref{eroTrans}} allows us to translate the structuring element $B$ by $q = (rx^B_{\hat{m}}, y^B_{\hat{m}})$. So $o \in B_{-q}$ and $(R^B_{\hat{m}})_{-q} = \langle lx_{\hat{m}}^B - rx_{\hat{m}}^B, 0, 0 \rangle$. Hence the claim immediately follows by using \hyperref[Xcutlemma]{Lemma~\ref{Xcutlemma}}.
        \end{pf}

    Altough we can reduce our investigation to the points of $X_{\operatorname{cut}}$, we still need to evaluate $f^A_{X_{L_{\min}}}$ and $f^{A^t}_{X_{L_{\min}}}$. \hyperref[imgmista]{\figurename~\ref{imgmista}} demonstrates the efficiency of this theorem.		
    \section{Dilation Using RLE}
\label{ch:dilationUsingRLE}

	By taking use of the duality between erosion and dilation \cite{Soille2003}
	\begin{displaymath}
		\delta_B(X) = \Big[ \varepsilon_{B^t}(X^{\mathsf{c}}) \Big]^{\mathsf{c}},
	\end{displaymath}
	we are able to propose a fast dilation algorithm based on the previously developed erosion algorithm in a straightforward way. But we have to make some additional thoughts, because obviously $|X^{\mathsf{c}}| = \infty$ when $|X| < \infty$.

	Consider the rectangle $\operatorname{REC}^{l, r, t, b}$ with $l,r,t,b \in \mathds{Z}$ given by:
	\begin{displaymath}
		\operatorname{REC}^{l, r, t, b} := \{ (x,y) \in \mathds{Z}^2 \mid l \leq x \leq r, \enspace t \leq y \leq b \}.
	\end{displaymath}
	Then we denote the smallest rectangle which includes $X$ by $\operatorname{REC}^{\min}_X$.

	Obviously (by considering the definition of dilation \hyperref[def:dil]{(\ref{def:dil})}) there exists a sufficiently big rectangle $\operatorname{REC}^{\delta}$ such that $\delta_B(X)$ does not contain any points outside this rectangle:
	\begin{displaymath}
		\delta_B(X) \, \backslash \, \operatorname{REC}^{\delta} = \bigg( \Big[ \varepsilon_{B^t}(X^{\mathsf{c}}) \Big]^{\mathsf{c}} \bigg) \, \backslash \, \operatorname{REC}^{\delta} = \emptyset.
	\end{displaymath}
	Therefore we get:
	\begin{displaymath}
		\Big[ \varepsilon_{B^t}(X^{\mathsf{c}}) \Big]^{\mathsf{c}} \, \cap \, \operatorname{REC}^{\delta} = \Big[ \varepsilon_{B^t}(X^{\mathsf{c}}) \Big]^{\mathsf{c}}.
	\end{displaymath}
	Because of $|\operatorname{REC}^{\delta}| < \infty$ we are also allowed to
restrict $X^{\mathsf{c}}$ by a rectangle $\operatorname{REC}^{\varepsilon}$, such that
	\begin{equation}
		\Big[ \varepsilon_{B^t}(X^{\mathsf{c}} \cap \operatorname{REC}^{\varepsilon} ) \Big]^{\mathsf{c}} \cap \operatorname{REC}^{\delta} = \Big[ \varepsilon_{B^t}(X^{\mathsf{c}}) \Big]^{\mathsf{c}}.\label{dil:1}
	\end{equation}
    Still we have to find suitable sizes for the rectangles $\operatorname{REC}^{\varepsilon}$ and $\operatorname{REC}^{\delta}$. Our goal is to make them as small as possible while being able to compute their sizes efficiently. \\
	Because of \hyperref[eroTrans]{Theorem~\ref{eroTrans}} we can assume that $o \in B$. If this is not the case, we are allowed to translate $B$ by a suitable vector $q$ such that $o \in B_q$. Since
	\begin{displaymath}
		\delta_B(X) = \{ p \in \mathds{Z}^2 \mid (B^t)_p \cap X \neq \emptyset \}
	\end{displaymath}
	one can easily see, that a suitable rectangle $\operatorname{REC}^{\delta}$ is received by adding a border to $\operatorname{REC}^{\min}_X$ which has the same width as the structuring element on the left and right hand sides and the same height as the structuring element on the top and bottom. Now we only need to find a rectangle $\operatorname{REC}^{\varepsilon}$ such that \eqref{dil:1} holds. Therefore we just double the border of $\operatorname{REC}^{\delta}$ to obtain $\operatorname{REC}^{\varepsilon}$.
	\\
 With these thoughts we are able to reformulate the dilation algorithm using the proposed erosion algorithm:
	\begin{enumerate}
		\item We receive run-length encoded images $X$ and $B$ as input.
		\item\label{en:2} We calculate $B^t$ and the compact run-length representation of $X^{\mathsf{c}} \cap \operatorname{REC}^{\varepsilon}$.
		\item These images can be used to retrieve $\varepsilon_{B^t}(X^{\mathsf{c}} \cap \operatorname{REC}^{\varepsilon} )$ by executing \hyperref[alg:GenerateDistanceTransformX2]{\texttt{GenErosionTransX2cut}($X^{\mathsf{c}} \cap \operatorname{REC}^{\varepsilon}, B^t$)}.
		\item\label{en:4} In the last step we take the complement of the erosion and restrict the image to the size of the rectangle $\operatorname{REC}^{\delta}$:
					\begin{displaymath}
						\delta_B(X) = \Big[ \varepsilon_{B^t}(X^{\mathsf{c}} \cap \operatorname{REC}^{\varepsilon} ) \Big]^{\mathsf{c}} \cap \operatorname{REC}^{\delta}.
					\end{displaymath}
	\end{enumerate}
Please note that \ref{en:2}) and \ref{en:4}) can also be done on the fly as described in \hyperref[sec:dilationalgorithm]{Section~\ref{sec:dilationalgorithm}}.
	
   	\section{Proposed Algorithms}
\label{sec:proposedAlgorithms}

In this section we are going to describe fast erosion and dilation algorithms. Reference~\cite{Ehrensperger2012} provides several figures that demonstrate the principles of these algorithms.

\subsection{Proposed Erosion Algorithm}
\label{sec:erosionalgorithm}

Observation \eqref{def:ero2} and \hyperref[eroTrans]{Theorem~\ref{eroTrans}} allow us to focus our investigations at those pixels of $X$, which can be looked up very efficiently for RLE images. Due to \hyperref[Erosion:2]{Theorem~\ref{Erosion:2}}, we are allowed to drop all runs of $X$ during preprocessing which are shorter than the shortest run of $B$. \hyperref[Xcuttheorem]{Theorem~\ref{Xcuttheorem}} states that it is enough to investigate the pixels of $X_{\operatorname{cut}}$, which is obtained by removing the first $L^B_{\max} - 1$ pixels of every run of $X$, where $L^B_{\max}$ denotes the length of the longest run of $B$. Of course, runs that are shorter than $L^B_{\max}$ vanish completely.
In case of a miss, the \hyperref[Machado2009:9]{Jump-Miss Theorem} enables us to skip the analysis of certain pixels and whenever a hit occurs, the full eroded run can immediately be added by applying the \hyperref[Erosion:3]{Jump-Hit Theorem}.

A pseudocode that implements all of these methods is given by the Algorithms~\ref{alg:GenerateSkeletonB}, \ref{alg:GenerateDistanceTransformX2}, and \ref{alg:GetErosion2}.

					\begin{algorithm}[!t]
						\caption{GenerateSkeletonB}
						\label{alg:GenerateSkeletonB}
						\begin{algorithmic}[1]
							\begin{spacing}{1.4}
                                    {\fontsize{8}{8} \selectfont
									\Function{GenerateSkeletonB}{$B$}\Comment{(01)}
									\State $S^A_B \gets \emptyset$
									\State $L^B_{\min} \gets \infty$
									\ForAll{$R^B_m \subseteq B$}\Comment{(02)}
											\State $S^A_B \gets S^A_B \cup \left\{(rx^B_m, y^B_m)\right\}$\label{alg:genskelB:1}\Comment{(03)}
											\State $f^A_B((rx^B_m, y^B_m)) \gets |R^B_m|$\Comment{(04)}
											\State $L^B_{\min} \gets \min \left\{ L^B_{\min}, |R^B_m| \right\}$\Comment{(05)}
									\EndFor
								\State \textbf{return} ($\operatorname{List}(S^A_B), f^A_B, L^B_{\min}$)\Comment{(06)}
								\EndFunction
								\Statex{ }
								\Statex{comments:}
								\Statex{(01): returns $\operatorname{List}(S^A_B)$, $f^A_B$ and $L^B_{\min}$}
								\Statex{(02): $R^B_m$ is defined as $R^B_m = \, \langle lx^B_m, rx^B_m, y^B_m \rangle$}
								\Statex{(03): adds the rightmost element to $S^A_B$}
								\Statex{(04): erosion transform of this element equals the length of the}
                                \Statex{\qquad \ run}
								\Statex{(05): keeps track of the shortest run within $B$}
								\Statex{(06): $\operatorname{List}(S^A_B)$ means that $S^A_B$ is stored as a list}
                                \Statex
                                }
							\end{spacing}
						\end{algorithmic}
                        \vspace{-4mm}
					\end{algorithm}

					\begin{algorithm}[!t]
						\caption{GenErosionTransX2cut}
						\label{alg:GenerateDistanceTransformX2}
						\begin{algorithmic}[1]
							\begin{spacing}{1.4}
                                {\fontsize{8}{8} \selectfont
								\Function{GenErosionTransX2cut}{$X, L^B_{\min}, L^B_{\max}$}\Comment{(01)}
									\ForAll{$z \in M_X$}\Comment{(02)}
											\State $f^A_{X_{L_{\min}}}(z) \gets 0$
											\State $f^{A^t}_{X_{L_{\min}}}(z) \gets 0$
									\EndFor
									\State $X_{\operatorname{cut}} \gets \emptyset$
									\ForAll{$R^X_n \subseteq X$}\Comment{(03)}
											\If{$|R^X_n| \geq L^B_{\min}$}\Comment{(04)}
												\State $X_{\operatorname{cut}} \gets X_{\operatorname{cut}} \cup \langle lx^X_n + L^B_{\max} - 1, rx^X_n, y^X_n \rangle$\Comment{(05)}
												\State $j \gets 1$
												\For{$i \gets lx^X_n$ to $rx^X_n$}\Comment{(06)}
													\State $f^A_{X_{L_{\min}}}((i,y^X_n)) \gets j$
													\State $f^{A^t}_{X_{L_{\min}}}((i,y^X_n)) \gets |R^X_n| - j + 1$
													\State $j \gets j + 1$
												\EndFor
											\EndIf
									\EndFor
									\State \textbf{return} ($f^A_{X_{L_{\min}}}, f^{A^t}_{X_{L_{\min}}}, X_{\operatorname{cut}}$)
								\EndFunction
								\Statex{ }
								\Statex{comments:}
								\Statex{(01): returns $f^A_{X_{L_{\min}}}$, $f^{A^t}_{X_{L_{\min}}}$ and $X_{\operatorname{cut}}$}
								\Statex{(02): initializes $f^A_{X_{L_{\min}}}$ and $f^{A^t}_{X_{L_{\min}}}$; $M_X$ denotes a}
								\Statex{\qquad \ $k \times l$ array where $k := \text{width}(X)$ and $l := \text{height}(X)$}
								\Statex{(03): visits every run in $X$; $R^X_n$ is defined as $R^X_n = $}
                                \Statex{\qquad \ $\langle lx^X_n, rx^X_n, y^X_n \rangle$}
								\Statex{(04): only considers runs with length $\geq L^B_{\min}$}
								\Statex{(05): where $\langle a, b, c \rangle := \emptyset$ when $a > b$}
								\Statex{(06): visits every point within the given run and calculates the}
								\Statex{\qquad \  according erosion transform values}
                                \Statex
                                }
							\end{spacing}
						\end{algorithmic}
                        \vspace{-4mm}
					\end{algorithm}	
                    \begin{algorithm*}[!t]
                    \caption{GetErosion2cut}
					\label{alg:GetErosion2}
                    \vspace{2mm}
                    \begin{minipage}{0.6\textwidth}
    						\begin{algorithmic}[1]
    							\begin{spacing}{1.4}
                                    {\fontsize{8.5}{8.5} \selectfont
    								\Function{GetErosion2cut}{$X, B$}\Comment{(01)}
    									\State $L^B_{\max} \gets 0$
                                        \ForAll{$R^B_m \subseteq B$}\Comment{(02)}
                                            \If{$L^B_{\max} < |R^B_m|$}
                                                \State $L^B_{\max} \gets |R^B_m|$
                                                \State $q \gets (rx^B_m, y^B_m)$ \Comment{(03)}
                                            \EndIf
                                        \EndFor
    									\State $B_{-q} \gets \text{TranslateImage}(B,-q)$ \Comment{(04)}
    									\State $\operatorname{List}(S^A_{B_{-q}}), f^A_{B_{-q}}, L^{B}_{\min} \gets \text{GenerateSkeletonB}(B_{-q})$
    									\State $f^A_{X_{L_{\min}}}, f^{A^t}_{X_{L_{\min}}}, X_{\operatorname{cut}} \gets \text{GenErosionTransX2cut}(X, L^{B}_{\min}, L^{B}_{\max})$
    									\State $\varepsilon_{B_{-q}}(X) \gets \emptyset$
    									\ForAll{$R^X_n \subseteq X_{\operatorname{cut}}$}\Comment{(05)}
    											\State $x \gets lx^X_n$\Comment{(06)}
    											\While{$x \leq rx^X_n$}\Comment{(07)}
    												\State $\text{miss} \gets \text{false}$\Comment{(08)}
    												\State $s \gets \operatorname{head}(\operatorname{List}(S^A_{B_{-q}}))$\Comment{(09)}
    												\While{$((\text{not}(\text{miss})) \text{ AND } (\text{exists}(\text{next}(\operatorname{List}(S^A_{B_{-q}})))))$}\Comment{(10)}
    													\State $s \gets \text{next}(\operatorname{List}(S^A_{B_{-q}}))$\Comment{(11)}
    													\State $\text{Diff} \gets f^A_{B_{-q}}(s) - f^A_{X_{L_{\min}}}(s+(x,y^X_n))$
    													\While{(($\text{Diff} > 0$) AND ($x \leq rx^X_n$))}\Comment{(12)}
    														\State $x \gets x + \text{Diff}$\Comment{(13)}
    														\State $\text{Diff} \gets f^A_{B_{-q}}(s) - f^A_{X_{L_{\min}}}(s+(x,y^X_n))$
    														\State $\text{miss} \gets \text{true}$\Comment{(14)}
    													\EndWhile
    												\EndWhile		
    												\If{($\text{not}(\text{miss})$)}\Comment{(15)}
    													\State minDist $\gets \infty$
    													\ForAll{$s \in S^A_{B_{-q}}$}\Comment{(14)}
    															\State $\text{minDist} \gets \min (\text{minDist}, f^{A^t}_{X_{L_{\min}}}(s+(x,y^X_n)))$\Comment{(16)}
    													\EndFor
    													\State $\varepsilon_{B_{-q}}(X) \gets \varepsilon_{B_{-q}}(X) \cup \left<x, x + \text{minDist} - 1, y^X_n\right>$\Comment{(17)}
    													\State $x \gets x + \text{minDist} + 1$\Comment{(18)}
    												\EndIf
    											\EndWhile
    									\EndFor
    									\State $\varepsilon_B(X) \gets \text{TranslateImage}(\varepsilon_{B_{-q}}(X), -q)$\Comment{(19)}
    									\State{\textbf{return} ($\varepsilon_B(X)$)}
                                    \EndFunction
                                    \Statex
                                    }
    								\algstore{GetErosion2}
    							\end{spacing}
    						\end{algorithmic}
                    \end{minipage}
                    \hfill
                    \begin{minipage}{0.38\textwidth}
    						\begin{algorithmic}[1]
    							\begin{spacing}{1.63}
                                    {\fontsize{8}{8} \selectfont
    								\algrestore{GetErosion2}
    								\Statex{(01): returns $\varepsilon_B(X)$}
    								\Statex{(02): determines the length and the position}
                                    \Statex{\qquad \ of the rightmost pixel of the longest}
                                    \Statex{\qquad \ run within $B$}
    								\Statex{(03): sets $q$ to the rightmost pixel of the}
                                    \Statex{\qquad \ longest run}
    								\Statex{(04): $B_{-q}$ contains the origin, which is the}
                                    \Statex{\qquad \ rightmost pixel of the longest run of $B$}		
    								\Statex{(05): visits every run of $X_{\operatorname{cut}}$; $R^X_n$ is defined}
                                    \Statex{\qquad \ as $R^X_n = \, \langle lx^X_n, rx^X_n, y^X_n \rangle$}
    								\Statex{(06): stores the $x$-coordinate of point $h$: we}
                                    \Statex{\qquad \ are checking if $B_{h-q} \subseteq X$}
    								\Statex{(07): breaks the loop once our $x$ lies outside}
                                    \Statex{\qquad \ the given sequence}
    								\Statex{(08): variable needed for breaking the loop}
                                    \Statex{\qquad \ in case of a miss according to the}
                                    \Statex{\qquad \ \hyperref[Machado2009:9]{Jump-Miss Theorem}}
    								\Statex{(09): $s$ points at the head of $\operatorname{List}(S^A_{B_{-q}})$}
    								\Statex{(10): loop will be continued as long as there}
                                    \Statex{\qquad \ is no miss and there are more $s \in S^A_{B_{-q}}$}
                                    \Statex{\qquad \ which have not been visited yet}
    								\Statex{(11): gets the next element within}
                                    \Statex{\qquad \ $\operatorname{List}(S^A_{B_{-q}})$}
    								\Statex{(12): $B_{h-q} \nsubseteq X$}
    								\Statex{(13): jump by Diff number of pixels}
    								\Statex{(14): breaks the outer loop since $B_{h-q} \nsubseteq X$}
    								\Statex{(15): $B_{h-q} \subseteq X$, so the \hyperref[Erosion:3]{Jump-Hit Theorem}}
                                    \Statex{\qquad \ can be applied}
    								\Statex{(16): determines the minimum distance to}
                                    \Statex{\qquad \ the right to be able to apply the}
                                    \Statex{\qquad \ \hyperref[Erosion:3]{Jump-Hit Theorem}}
    								\Statex{(17): \hyperref[Erosion:3]{Jump-Hit Theorem}}
    								\Statex{(18): next point $(h-q)$ to look at will be at}
                                    \Statex{\qquad \ $x$-coordinate $(x + \text{minDist} + 1)$}
    								\Statex{(19): we retrieve $\varepsilon_B(X)$ by applying}
                                    \Statex{\qquad \ \hyperref[eroTrans]{Theorem~\ref{eroTrans}}}
                                    \Statex
                                    }
    							\end{spacing}
    						\end{algorithmic}
                    \end{minipage}
                    \vspace{-4mm}
                \end{algorithm*}

\subsection{Proposed Dilation Algorithm}
\label{sec:dilationalgorithm}

By adapting the described erosion algorithm, all required operations (calculating the complement, intersecting with rectangles) can be done on the fly. Recall that we obtained the following formula in \hyperref[ch:dilationUsingRLE]{Section~\ref{ch:dilationUsingRLE}}:
\begin{displaymath}
    \delta_B(X) = \Big[ \varepsilon_{B^t}(X^{\mathsf{c}} \cap \operatorname{REC}^{\varepsilon} ) \Big]^{\mathsf{c}} \cap \operatorname{REC}^{\delta}.
\end{displaymath}
By replacing line~\ref{alg:genskelB:1} in \hyperref[alg:GenerateSkeletonB]{\texttt{GenerateSkeletonB}} with $S^A_B \gets S^A_B \cup \left\{(-lx^B_m, -y^B_m)\right\}$ we are able to generate the skeleton of $B^t$ on the fly. In \hyperref[alg:GenerateDistanceTransformX2]{\texttt{GenErosionTransX2cut}} we immediately generate the erosion transform of $X^\mathsf{c} \cap \operatorname{REC}^{\varepsilon}$ by adding the sequences between two runs $R^X_n \subseteq X$ instead of the runs themselves. Additionally we need to add the border with width and height of $B$. Last but not least, we can modify \hyperref[alg:GetErosion2]{\texttt{GetErosion2cut}} by adding the misses instead of the hits to the dilated image.

   	\section{Runtime Analysis}
\label{ch:runtimeanalysis}

\subsection{Erosion}

In \hyperref[subsec:JMT]{Section~\ref{subsec:JMT}} we stated that the runtime complexity of the actual erosion algorithm is $\mathcal{O}(|X||S^A_B|)$. Next we are going to investigate the preprocessing steps. To build the skeleton $S^A_B$ we need to visit every run and store the length of the run as well as the rightmost pixel. Then the complexity is given by $\mathcal{O}(M)$, where $M$ denotes the number of runs of $B$ and obviously this leads to $|S^A_B| = M$. To obtain $f^A_{X_{L_{\min}}}$ and $f^{A^t}_{X_{L_{\min}}}$ we need to visit every point of $X$, thus its complexity is given by $\mathcal{O}(|X|)$. Further $X_{\operatorname{cut}}$ is built while determining $f^A_{X_{L_{\min}}}$ and $f^{A^t}_{X_{L_{\min}}}$ and does not affect the given complexity. Putting all together, we get a complexity of
\begin{equation}
    \mathcal{O}(|X|M + M + |X|) = \mathcal{O}(|X|M).\label{ero:complexity}
\end{equation}

\subsection{Dilation}

Since this algorithm is taking use of the duality between erosion and dilation, and building the complements is done on the fly, we get the following complexity (compare with \eqref{ero:complexity}):
\begin{displaymath}
    \mathcal{O}(|X^{\mathsf{c}} \cap \operatorname{REC}^{\varepsilon}| \cdot M).
\end{displaymath}

   	\section{Experimental Results}
\label{ch:experimentalResults}
\begin{figure}[!t]
        \centering
		\includegraphics[width=0.45\textwidth]{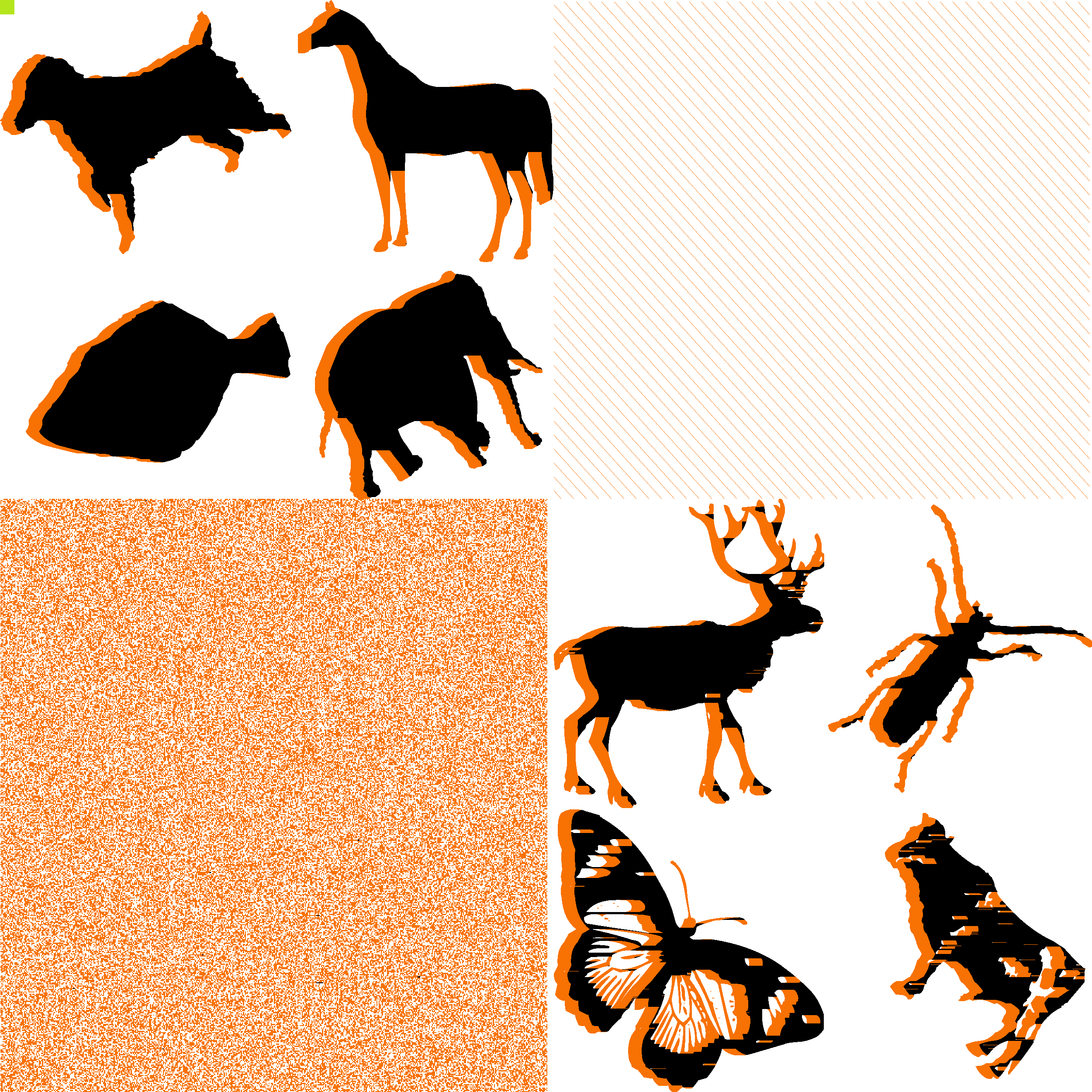}
		\caption{The size of this image is $2500 \times 2500$ pixels. The gray square (green in the online version) in the upper-left corner represents the structuring element, the gray parts (orange in the online version) of the image demonstrate the efficiency of \hyperref[Xcuttheorem]{Theorem~\ref{Xcuttheorem}}. This theorem states that the investigation of the gray pixels can be skipped while calculating the erosion. The source of the original image is \url{http://www.imageprocessingplace.com/}.}
		\label{imgmista}
	\end{figure}

In this section we are going to compare the runtimes of the proposed algorithms. Implementations of these algorithms are available in \texttt{C++} and can be found at \url{https://numerical-analysis.uibk.ac.at/g.ehrensperger}. We are comparing them with the implementations of the free library \textit{OpenCV} (developed by Intel) v2.4.2 and with the algorithms proposed in \cite{Kim2005} and \cite{Machado2009}. The latter two got implemented by Machado and the source code is available at \url{http://score.ime.usp.br/~dandy/mestrado.php}. This implementations are also used in our tests. Additionally he implemented various variants of his own algorithm. In the following plots we took the pointwise minimum of the runtimes of his algorithms. The conversion of $X$ and $B$ into the used input format of the various algorithms is not part of the given runtimes.\footnote{In \cite{Machado2009} the conversion of the raster graphics $X$ and $B$ into the used input format of the algorithms was part of the given runtimes.} \\
All tests were carried out on a workstation with Intel Core i7-3770K (for the following tests only one core was used), 32 GB DDR3-1333 RAM and OS \textit{Ubuntu 13.04 64 Bit}. CPU-stepping, overclocking settings, and various
energy saving options had been disabled. The test environment was compiled with \textit{GCC (the GNU Compiler Collection)} v4.7.3 (official project URL: \url{http://gcc.gnu.org/}) and compiler flag \texttt{O2}. The following results are the arithmetic mean values of three iterations. We used the image in \hyperref[imgmista]{\figurename~\ref{imgmista}} to compare the runtimes of the algorithms listed in  \hyperref[tab:Algorithmen]{Table~\ref{tab:Algorithmen}}.
	\begin{table}[!t]
        \renewcommand{\arraystretch}{1.3}
		\caption{Abbreviations of the tested algorithms}
		\label{tab:Algorithmen}
		\centering
		\begin{tabular}{c||c}
			\hline \textbf{\textsc{algorithm}}			& \textbf{\textsc{source}} \\
			\hline \hline \textbf{RLE} 					& erosion as in \cite{Kim2005} \\
			\hline \textbf{Machado}						& erosion as in \cite{Machado2009} \\
			\hline \textbf{OpenCV}						& erosion as in \cite{opencv}	 \\
			\hline \textbf{eEJMH}						& erosion algorithm as in \hyperref[sec:erosionalgorithm]{Sec.~\ref{sec:erosionalgorithm}} \\
			\hline \textbf{dRLE}							& dilation as in \cite{Kim2005}	\\
			\hline \textbf{dOpenCV}						& dilation as in \cite{opencv} \\
			\hline \textbf{dEJMH}						& dilation algorithm as in \hyperref[sec:dilationalgorithm]{Sec.~\ref{sec:dilationalgorithm}} \\
			\hline
		\end{tabular}
	\end{table}	

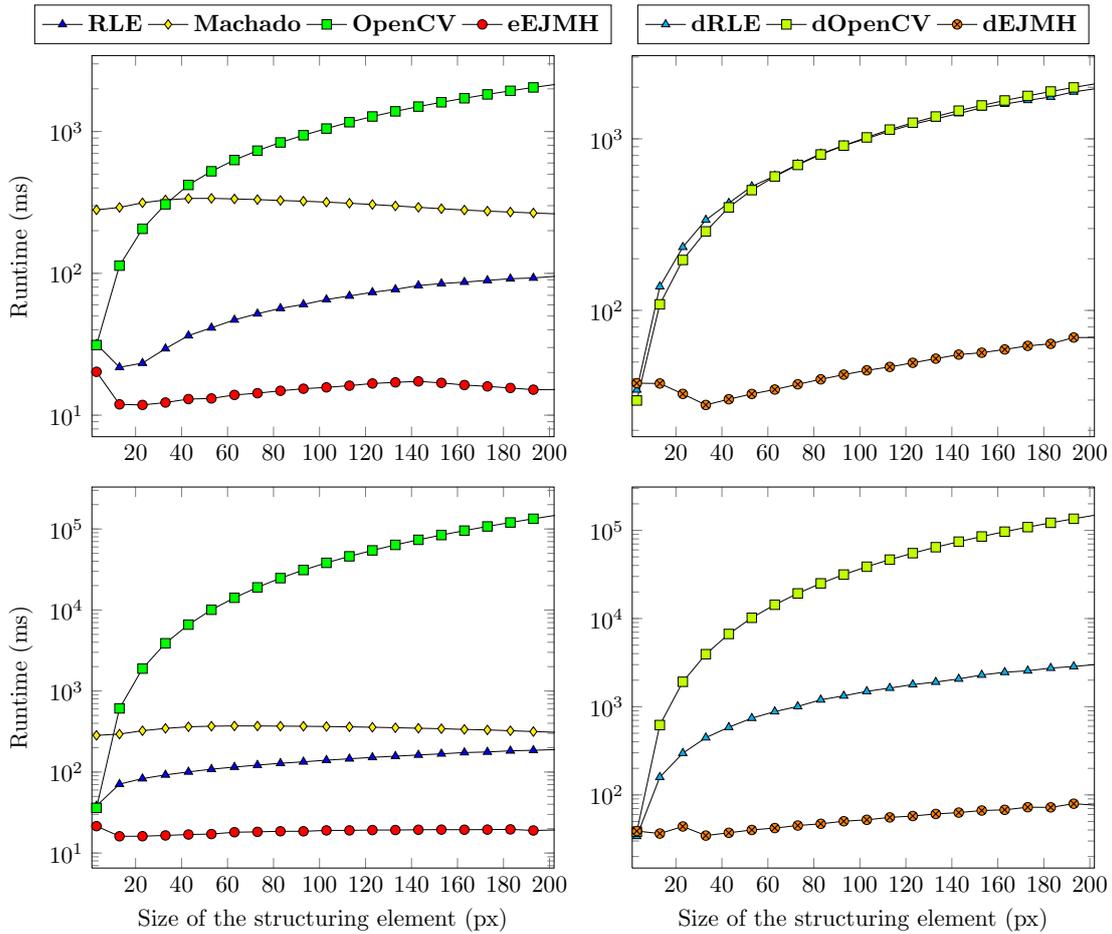
\begin{figure*}[!t]
        \centering
        \resizebox{0.9\textwidth}{!}{
            \begin{minipage}{\textwidth}
                \centering		
                \begin{minipage}{0.48\textwidth}								
        			\begin{tikzpicture}
                        \pgfplotsset{every axis legend/.append style={at={(0.5,1.03)},anchor=south}}
                        \pgfplotsset{try min ticks=10}
        				\begin{semilogyaxis}[legend cell align={left},
        									 xlabel={~},
                                             xmin=1,
                                             xmax=202,
        									 ylabel={Runtime (ms)},
        									 y tick label style={/pgf/number format/1000 sep=},
                                             legend columns=4]
        					\pgfplotstableread{Runtime_Analysis/Erosion_square_mista.dat}
        					\datatable
        					\addplot[mark=triangle*, color=blue, draw=black] table[y = RLE] from \datatable;
        					\addplot[mark=diamond*, color=yellow, draw=black] table[y = JmDvM1] from \datatable;
        					\addplot[mark=square*, color=green, draw=black] table[y = openCV] from \datatable;
        					\addplot[mark=*, color=red, draw=black] table[y = Ehren2cut] from \datatable;
        					\legend{%
        						\textbf{RLE},
        						\textbf{Machado},
        						\textbf{OpenCV},
        						\textbf{eEJMH},
        					}
        				\end{semilogyaxis}
        			\end{tikzpicture}
                \end{minipage}
                \quad
                \begin{minipage}{0.48\textwidth}		
        			\begin{tikzpicture}
                        \pgfplotsset{every axis legend/.append style={at={(0.5,1.03)},anchor=south}}
                        \pgfplotsset{try min ticks=10}
        				\begin{semilogyaxis}[legend cell align={left},
        									 xlabel={~},
                                             xmin=1,
                                             xmax=202,
        									 ylabel={~},
        									 y tick label style={/pgf/number format/1000 sep=},
                                             legend columns=3]
        					\pgfplotstableread{Runtime_Analysis/Dilation_square_mista.dat}
        					\datatable
        					\addplot[mark=triangle*, color=cyan, draw=black] table[y = dRLE] from \datatable;
        					\addplot[mark=square*, color=lime, draw=black] table[y = dopenCV] from \datatable;
        					\addplot[mark=otimes*, color=orange, draw=black] table[y = dEhrencut] from \datatable;
        					\legend{%
        						\textbf{dRLE},
        						\textbf{dOpenCV},
        						\textbf{dEJMH},
        					}
        				\end{semilogyaxis}
        			\end{tikzpicture}
                \end{minipage} \\
                \begin{minipage}{0.48\textwidth}					
                    \begin{tikzpicture}
                        \pgfplotsset{try min ticks=10}
        				\begin{semilogyaxis}[xlabel={Size of the structuring element (px)},
                                             xmin=1,
                                             xmax=202,
        									 ylabel={Runtime (ms)},
        									 y tick label style={/pgf/number format/1000 sep=}]
        					\pgfplotstableread{Runtime_Analysis/Erosion_diamond_mista.dat}
        					\datatable
        					\addplot[mark=triangle*, color=blue, draw=black] table[y = RLE] from \datatable;
        					\addplot[mark=diamond*, color=yellow, draw=black] table[y = JmDvM1] from \datatable;
        					\addplot[mark=square*, color=green, draw=black] table[y = openCV] from \datatable;
        					\addplot[mark=*, color=red, draw=black] table[y = Ehren2cut] from \datatable;
        				\end{semilogyaxis}
        			\end{tikzpicture}
                \end{minipage}
                \quad
                \begin{minipage}{0.48\textwidth}			
        			\begin{tikzpicture}
                        \pgfplotsset{try min ticks=10}
        				\begin{semilogyaxis}[xlabel={Size of the structuring element (px)},
                                             xmin=1,
                                             xmax=202,
        									 ylabel={~},
        									 y tick label style={/pgf/number format/1000 sep=}]
        					\pgfplotstableread{Runtime_Analysis/Dilation_diamond_mista.dat}
        					\datatable
        					\addplot[mark=triangle*, color=cyan, draw=black] table[y = dRLE] from \datatable;
        					\addplot[mark=square*, color=lime, draw=black] table[y = dopenCV] from \datatable;
        					\addplot[mark=otimes*, color=orange, draw=black] table[y = dEhrencut] from \datatable;
        				\end{semilogyaxis}
        			\end{tikzpicture}
                \end{minipage}
            \end{minipage}
        }
        \caption{Upper left: Erosion with a square-shaped structuring element. Upper right: Dilation with a square-shaped structuring element. Lower left: Erosion with a diamond-shaped structuring element. Lower right: Dilation with a diamond-shaped structuring element.}
        \label{fig:plot}
    \end{figure*}

    \noindent The results are given in \hyperref[fig:plot]{\figurename~\ref{fig:plot}}. Note that the plots' $y-$axes are scaled logarithmically.
    As can be seen, the algorithms proposed in this paper tremendously improve the runtime of both, the erosion and dilation operator, over the compared algorithms on the test image given in \hyperref[imgmista]{\figurename~\ref{imgmista}}. We observe that the execution times of \textbf{RLE}, \textbf{dRLE}, \textbf{OpenCV}, and \textbf{dOpenCV} grow with the size of the structuring element. In contrary \textbf{eEJMH} and \textbf{Machado} benefit from the size of the structuring element. This can be explained, since the bigger the structuring element the larger the jumps allowed by the \hyperref[Machado2009:9]{Jump-Miss Theorem}. The clear advantage of \mbox{\textbf{eEJMH}} over \textbf{Machado} results from the various new methods described in \hyperref[ch:erosionUsingRLE]{Section~\ref{ch:erosionUsingRLE}}. Although \mbox{\textbf{dEJMH}} is based on the same methods as \mbox{\textbf{eEJMH}}, we observe that the runtime still grows with the size of the structuring element. This is because we work with the inverted input image $X^{\mathsf{c}} \cap \operatorname{REC}^{\varepsilon}$ and the rectangle $\operatorname{REC}^{\varepsilon}$ depends on the size of the structuring element. Still \textbf{eEJMH} and \textbf{dEJMH} seem to provide a better alternative in most cases.

    \section{Conclusions}
\label{ch:conclusions}
In this paper we developed new ideas to speed up the calculation of erosion and dilation. We also proposed fast algorithms to perform these operations with arbitrary structuring elements. Further we determined the runtime complexity of our algorithms which is given by $\mathcal{O}(|X|M)$ for erosion and by $\mathcal{O}(|X^{\mathsf{c}} \cap \operatorname{REC}^{\varepsilon}| \cdot M)$ for dilation, where $X$ is the input image, $B$ the structuring element, $M$ denotes the number of runs of $B$, and $\operatorname{REC}^{\varepsilon}$ is a rectangle of the size of $X$ plus a border that is twice the size of $B$.  Finally experiments confirmed that our algorithms provide huge speedup compared to some other well known implementations.

	\section*{Acknowledgments}
	\label{ch:acknowledgements}
	This work was carried out in collaboration with Besi Austria GmbH (former Datacon Technology GmbH).

    \bibliographystyle{elsarticle-num}
    \bibliography{./Bib/ref}
	
	\vspace{3em}
	
\noindent \textbf{Alexander Ostermann}
received the Mag. and Dr. degrees in science (mathematics and numerical analysis) from the University of Innsbruck, Austria, in 1984 and 1988, respectively.

He was a Faculty Member of the School of Civil Engineering with the University of Innsbruck from 1982 to 2002. From 1988 to 1990 he was a Post-Doctoral Researcher with the University of Geneva, Switzerland and from 1999 to 2000, he was Professor with the same university. Since 2002, he has been Full Professor with the University of Innsbruck. He was Head of the Department of Mathematics from 2005 to 2008 and Dean of the School of Mathematics, Computer Science and Physics from 2008 to 2013. Since 2013, he has been Head of the Center for Scientific Computing with the same university. His current research interests include numerical analysis, geometry, and mathematical methods in science and engineering. Together with Gerhard Wanner, he has written the book ``Geometry by Its History'' (Berlin: Springer, 2012).

Prof. Ostermann is board member of the Austrian Mathematical Society (\"OMG), and member of various mathematical societies (AMS, SIAM, EMS).\\
\\
\textbf{Felix Schwitzer}
received the Mag. and Dr. degrees in science (mathematics and numerical analysis) from the University of Innsbruck, Austria, in 1991 and 1999, respectively.

He is a software engineer at Besi Austria GmbH with focus on the development of computer vision algorithms and applications for automatization.\\
\\
\textbf{Gregor Ehrensperger}
received the B.Sc. degree in mathematics from the University of Hagen, Germany, in 2011 and the Dipl.-Ing. degree in technical mathematics at the University of Innsbruck, Austria, in 2014. Currently he is  pursuing the B.Sc. in computer science at the University of Innsbruck.
%    \twocolumn[\input{biographies}]

\end{document}